\makeatletter
\@ifundefined{caps}{}{}
\makeatother

\documentclass[letterpaper, 10 pt, conference]{ieeeconf}  

\IEEEoverridecommandlockouts                              

\usepackage{graphicx}
\usepackage{gensymb}
\usepackage{amsmath}
\usepackage{cite}
\title{\LARGE \bf A Matter of Height: The Impact of a Robotic Object on \\Human Compliance}

\author{Michael Faber, Andrey Grishko, Julian Waksberg, David Pardo, Tomer Leivy, Yuval Hazan, \\ Emanuel Talmansky, Benny Megidish, and Hadas Erel 
\thanks{*This work was supported by the Israeli Innovation Authority as part of the HRI Consortium.}
\thanks{Media Innovation lab, Reichman University, Herzliya. {\tt\small Michael.Faber@milab.runi.ac.il, hadas.erel@milab.runi.ac.il}}%
}

\begin{document}

\maketitle
\thispagestyle{empty}
\pagestyle{empty}

\begin{abstract}
Robots come in various forms and have different characteristics that may shape the interaction with them. In human-human interactions, height is a characteristic that shapes human dynamics, with taller people typically perceived as more persuasive. In this work, we aspired to evaluate if the same impact replicates in a human-robot interaction and specifically with a highly non-humanoid robotic object. The robot was designed with modules that could be easily added or removed, allowing us to change its height without altering other design features. To test the impact of the robot's height, we evaluated participants' compliance with its request to volunteer to perform a tedious task. In the experiment, participants performed a cognitive task on a computer, which was framed as the main experiment. When done, they were informed that the experiment was completed. While waiting to receive their credits, the robotic object, designed as a mobile robotic service table, entered the room, carrying a tablet that invited participants to complete a 300-question questionnaire voluntarily. We compared participants' compliance in two conditions: A Short robot composed of two modules and 95cm in height and a Tall robot consisting of three modules and 132cm in height. Our findings revealed higher compliance with the Short robot's request, demonstrating an opposite pattern to human dynamics. We conclude that while height has a substantial social impact on human-robot interactions, it follows a unique pattern of influence. Our findings suggest that designers cannot simply adopt and implement elements from human social dynamics to robots without testing them first.

\end{abstract}

\section{Introduction}
As robots increasingly enter our lives, mapping their impact becomes a primary challenge \cite{broadbent2017interactions, brondi2021we}. In addition to designing robots to serve a specific function \cite{goodrich2008human} and to be accepted by humans \cite{bishop2019social}, we also need to consider the social aspects that people automatically assign to interactions with them \cite{hoffman2014designing, erel2019robots, erel2021excluded, duffy2003anthropomorphism, novikova2014design}. Previous studies have examined how interacting with robots can lead to both positive \cite{tennent2019micbot, erel2022enhancing, adi2022non} and negative \cite{erel2021excluded, salomons2021minority, hitron2022ai} social effects, similar to those observed in human dynamics. Most of these studies focused on the impact of the robots' behavior and communication modalities (i.e., speech and gestures) while controlling for the robot's characteristics and design. However, one of the most significant advantages of robots as autonomous devices is the freedom in their design and the ability to adjust it to their specific function \cite{hoffman2014designing}. This suggests that robots are expected to employ varying design characteristics, including various forms, colors, sizes, and heights. These characteristics are already known to have a social impact in human-human interactions and human-social dynamics \cite{bartneck2018robots, bernotat2017shape, saerbeck2010perception, hiroi2008bigger}.

 \begin{figure}[t]
     \centering
     \vspace{0.2cm}
\includegraphics[width=1\linewidth]{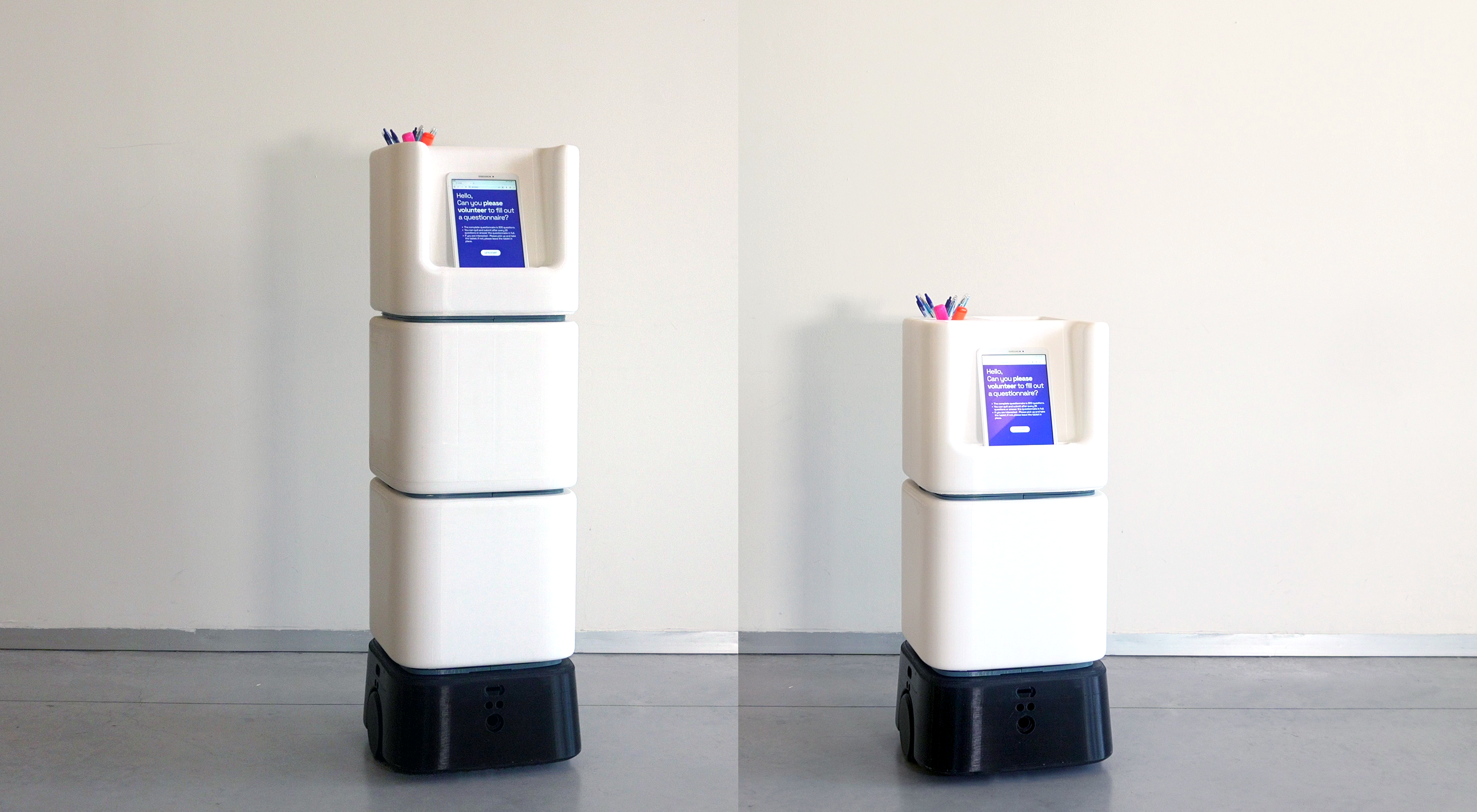}
     \caption{The modular robotic service table: Short and Tall versions.}
     \label{fig:MORPHY - A Modular Robotic Platform}
 \end{figure}

Previous work suggests that robot design characteristics may also significantly impact social dynamics when interacting with them \cite{bartneck2018robots, bernotat2017shape, saerbeck2010perception}. People respond differently to robots based on their color \cite{bartneck2018robots}, gendered design cues (feminine vs. masculine) \cite{bernotat2017shape}, and motion style \cite{saerbeck2010perception}. Another interesting design characteristic is the robot's height. Previous studies indicated that it has a diverse impact on the social dynamics between the human and the robot \cite{hiroi2008bigger}, with some reporting similar effects to those observed with human height. A common finding in human-human interactions is that taller people are perceived as more competent, dominant, charismatic, persuasive, and likely to occupy positions of leadership and authority \cite{young1996height, jaeger2011thing,stogdill1948personal,stulp2013tall}. As a result, people tend to comply with taller people's requests \cite{young1996height, higham1992rise, stogdill1948personal}. Similarly, studies evaluating interactions with robots have shown that robots' height impacts compliance with the robot's operator, with taller robots leading to increased compliance \cite{rae2013influence} and trust \cite{gervasi2024does}. However, other studies indicated that people prefer to collaborate with short robots \cite{samarakoon2022review, walters2009preferences} and perceive the interaction with them as safer \cite{joosse2021appearance}. This points to a potential preference for shorter robots—opposite to the height-related effects seen in human-human interactions (where being taller is associated with leadership and charisma \cite{judge2004effect}).
A factor that may mediate the impact of robotic height is the human-likeness of the robot \cite{walters2009preferences}. As suggested earlier, one of the greatest advantages of robots is the ability to adjust their design to their intended function \cite{hoffman2014designing}. Accordingly, robots will take many forms, ranging from robots with a humanoid appearance to familiar objects and abstract objects. The automatic tendency to anthropomorphize even highly non-humanoid robots \cite{erel2019robots} suggests that height may influence social dynamics in HRI, regardless of the robot’s degree of human-likeness. Nevertheless, it is unclear if this social impact would parallel the effect of height in human-human interactions \cite{erel2024rosi}. While it is reasonable to assume that interactions with humanoid robots would follow similar patterns to those with other humans (see, for example, \cite{maj2024can}), it has yet to be explored if a similar pattern would emerge when interacting with highly non-humanoid robots. The robot's machine-like nature may mediate the social influence of the robot's height, leading to different and unique effects \cite{joosse2021appearance}. 
 
In this work, we designed and developed a robotic object (i.e., a mobile service table) to evaluate the impact of the robot's height on participants' willingness to comply with its request. The robotic object was designed as a modular platform composed of various modules that could be easily added or removed to change the robot's height. Using the flexibility offered by the platform, we could change the robot's height without changing other design features (see Figure \ref{fig:MORPHY - A Modular Robotic Platform}). We compared: (1) an interaction with a Short robot, composed of two modules and 95 cm height; (2) and an interaction with a Tall robot, composed of three modules and 132 cm height. In the experiment, participants performed a cognitive task on a computer. When done, they were informed that the experiment was complete. While waiting to receive their credits, the robotic platform entered the room, carrying a tablet with a call for voluntarily filling out a 300-question-long questionnaire \cite{soto2017next}. To test if the impact of the robot's height would parallel the effect of height in human-human interactions, we evaluated the compliance with the robot's request (i.e., the number of questions participants volunteered to answer) and the robot's perception.

 \begin{figure*}[h]
     \centering
      \vspace{0.3cm}
\includegraphics[width=0.85\linewidth]{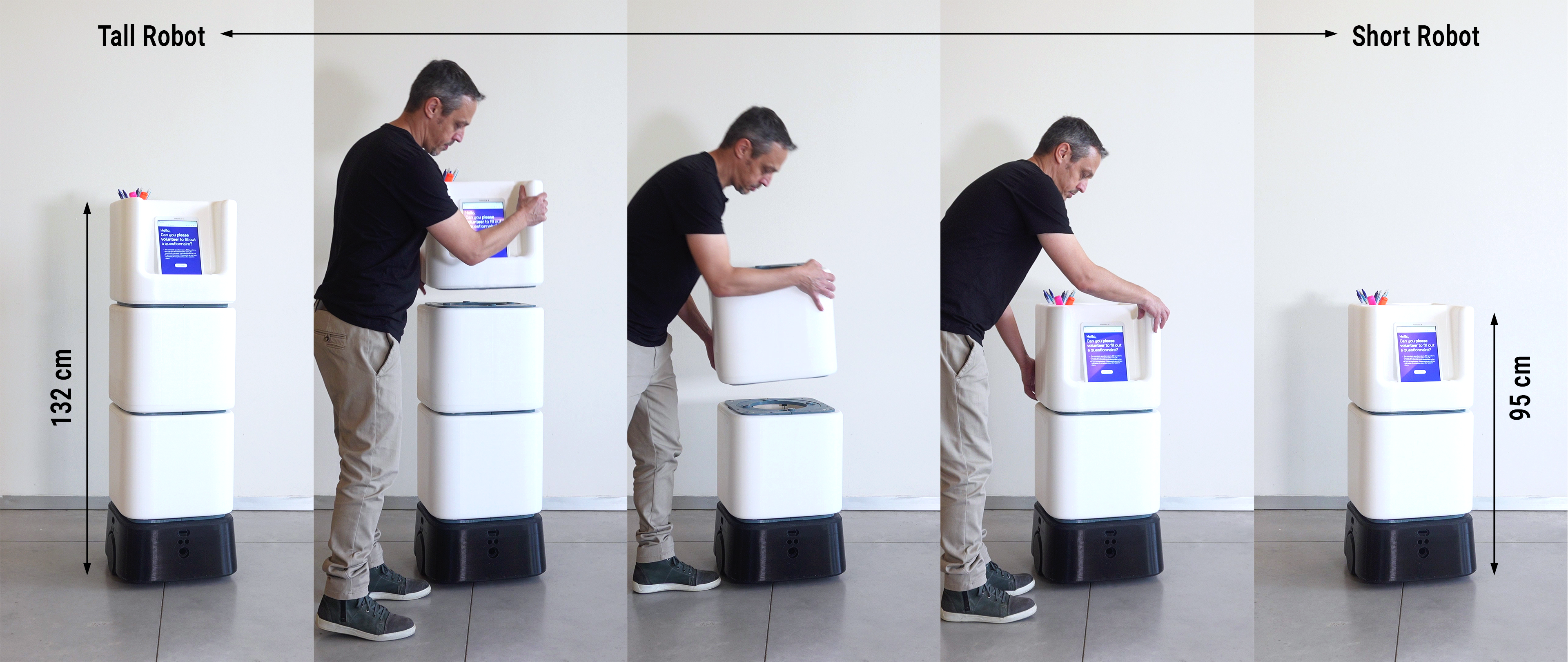}
     \caption{From Tall to Short, and vice versa. Robot assembly.}
     \label{fig:robotfigure}
 \end{figure*}


\section{Related Work}
Relevant previous studies include HRI studies that explored the impact of a robot's height and compliance with robots' requests. 

\subsection{The impact of a robot's height}

Previous studies in HRI have shown that a robot’s height can influence the interaction dynamics and the way the robot is perceived \cite{rae2013influence, hiroi2008bigger, hiroi2016height, walters2009preferences, joosse2021appearance, samarakoon2022review}. While some of these studies indicated effects similar to those observed in human-human interactions \cite{rae2013telepresence, hiroi2016height}, others showed opposite patterns \cite{joosse2021appearance, walters2009preferences}. Among the studies that indicated similar effects to those observed in human-human interactions, Rae et al. (2013) studied compliance to requests presented by operators of telepresence robots. They tested whether the robot's height would impact the perceived authority and persuasiveness of the person operating it. They found that when the telepresence robot was shorter than the participants, its operators' authority was lower, and the participants reported feeling more dominant in the conversation \cite{rae2013telepresence}. Similarly, Walters et al. (2009) found that taller humanoid robots were perceived as more conscientious, while short humanoid robots were viewed as more neurotic and less conscientious \cite{walters2009preferences}, replicating the tendency to attribute positive characteristics to taller individuals. 

Other studies suggested that height may have an opposite effect to that found in human-human interactions, with people showing clear preferences for short robots. For example, Hiroi and Ito (2016) found that participants preferred to converse with a short robot, specifically with a robot that did not reach their eye level \cite{hiroi2016height}. Similarly, Joosse et al. (2021) found that shorter robots were perceived as safer and more comfortable to interact with. This suggests that humans in HRI may prioritize emotional safety and comfort over authority, which contrasts with human interactions where height positively correlates with competence and status \cite{joosse2021appearance}.

These studies indicate that a robot's height can impact the interactions with humans, the robot's perception, and people's preferences. We extend this line of studies by focusing on the impact of a non-humanoid robot's height and its influence on participants' willingness to comply with the robot's request. We tested whether, as in human-human interactions, a taller robot would lead to increased compliance—or whether the specific case of a non-humanoid robot would result in a different pattern. 

\subsection{Compliance with robots' requests} 

Several studies have explored human tendencies to comply with robots' requests \cite{lee2016role, nielsen2022prosocial, erel2024power}. The robot’s emotional behavior \cite{kuhnlenz2018effect, shiomi2017hug}, its ability to align with socially appropriate behaviors \cite{connolly2020prompting}, its social role \cite{saunderson2021persuasive}, its persuasion strategy \cite{lee2019robotic}, and its appearance \cite{kim2014effect} were all shown to encourage compliance with its requests \cite{connolly2020prompting, martin2020investigating, zaga2017gotta}. For example, Moshkina (2012) indicated that when a robot expressed affect, particularly through nonverbal expressions of negative mood such as nervousness and fear, it enhanced participants’ compliance with evacuation requests, leading to earlier and faster responses \cite{moshkina2012improving}. Similarly, Wills et al. (2016) found that participants were more compliant with a donation request when it came from a robot that followed human-like social norms, demonstrating appropriate gaze and facial expression cues, compared to a robot that did not exhibit social responsiveness to its surroundings \cite{wills2016socially}. Another aspect of HRI that impacts people's willingness to comply with a robot's request is its social role. Robots framed as peers were shown to elicit more compliance than those framed as supervisors. For example, Saunderson and Nejat (2021) found that participants were more willing to follow a peer robot's requests in comparison to a robot framed as an authority figure, suggesting that social framing may also shape compliance to robots \cite{saunderson2021persuasive}.

These studies indicate that various factors influence people's tendency to comply with a robot's request. We further explored whether robotic design characteristics can impact social dynamics and enhance participants' compliance. Specifically, we evaluated the impact of height in interactions with a highly non-humanoid robot.

\vspace{0.5cm}
\section{Design}

To test the impact of a robotic object's height, we designed a modular robot that allows us to modify its height without changing other parameters, such as color, shape, and materials (see Figure \ref{fig:robotfigure}). Since our focus was on the specific case of a non-humanoid robotic object, we designed a robotic service table that would be assimilated within the lab context. The robotic platform consisted of several attachable modules that afforded extensive flexibility in assembly, similar to interlocking LEGO bricks, thereby enabling a range of configurations. The modules can be stacked on top of each other, allowing the robot to reach various heights while keeping all other aspects constant. The system was composed of three types of primary modules: (1) a mobile base module integrating the driving chassis, electronics, and batteries; (2) intermediary spacer modules that facilitated height variation; and (3) a top module designed to carry office equipment.

\subsection{Design principles}
During development, a multidisciplinary team—including a designer, an HRI researcher, an industrial designer, software developers, and a cognitive psychologist, established a set of design guidelines grounded in our research hypothesis. The core design principles were as follows:
\subsubsection{Flexibility} The platform must allow a rapid and seamless transition between different height configurations.
\subsubsection{Neutrality} The design aimed for a neutral, industry-standard appearance that minimized extraneous associations and could be related to a lab context.
\subsubsection{Affordability and Replicability} The system was designed to be cost-effective and reproducible, facilitating widespread adoption and further advancements within the HRI research community. Future enhancements and integration of additional modules should be straightforward.

\begin{figure}[b]
     \centering
\includegraphics[width=0.9\linewidth]{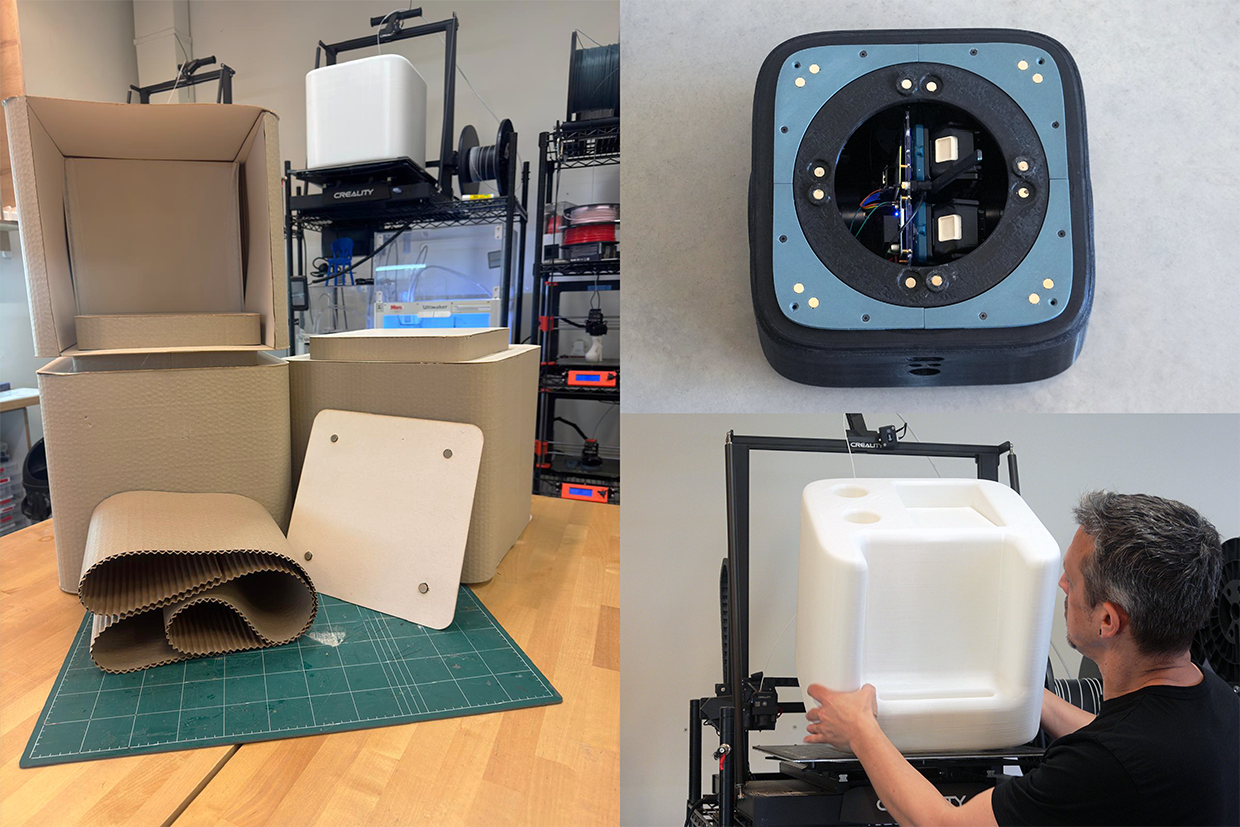}
     \caption{The design process.}
     \label{fig:Design1}
 \end{figure}

The initial design phase employed full-scale, low-fidelity cardboard mock-ups (see Figure \ref{fig:Design1}) to establish dimensionality and ergonomics. These prototypes were tested in a pilot study to ensure comfortable interaction and that critical elements, such as equipment accessibility, remained within easy reach while the participant was sitting, whether the platform was configured in its taller or shorter states.

\subsection{Advanced design phase}
In the subsequent phase, we refined the design with detailed attention to the chassis, enclosure, and modular components (see Figure \ref{fig:Design1}). A cubic form factor was selected over alternative geometries due to its alignment with the standard shape and dimensions of desks, offering superior functional clarity and optimal space utilization \cite{kim2019assembly}. Additionally, rounded corners were incorporated to evoke a more approachable and safer aesthetic compared to sharp edges \cite{bar2006humans, bar2007visual}. Designing the chassis to accommodate all electronic components, support the robot's weight, and achieve the required stability involved an intensive iteration process, that resulted in over four fully functional mid-fidelity prototypes. The process began with a combination of 3D-printed and laser-cut parts, transitioning to a 3D-printed design to enhance rigidity and minimize the required tools for manufacturing. 

The mobile module used two freewheel-mounted motorized wheel configurations, a common indoor robotic design that facilitated maneuverability and in-place turning. Each spacer module was designed with a central aperture along the platform’s core, providing the flexibility for future integration of additional electronics and features. To ensure secure stacking, all modules utilized a uniform magnetic connector system, achieving a robust assembly. The top module was developed through several iterative refinements, culminating in a design that supported convenient access.

\subsection{Future directions}
We envisioned the continued evolution of this platform as an open-source resource, with the integration of additional compatible modules and functionalities to further support HRI research in the field of mobile robotic platforms.

\subsection{Technical implementation}\label{sec:Technical implementation}
The robot's hardware consisted of an ESP32 micro-controller, two DC motors with wheels for mobility, and two Makita batteries (18V, 5Ah) providing the power supply through an adapter. The motor control was managed through an H-bridge motor driver, enabling both forward and reverse motion, with speed adjustments controlled via the controller’s triggers. The BluePad32 library was utilized to handle communication between the ESP32 micro-controller and the PS4 controller itself. Development and integration were carried out using PlatformIO, a cross platform tool for embedded development. In order to develop a versatile platform that could be easily controlled remotely and adapted to different tasks through a modular structure, we utilized the ESP32 micro-controller as the central control unit. The robot is operated using a Sony PlayStation 4 controller, providing an intuitive and responsive interface for real-time control that allows the robot to move seamlessly using a Wizard of Oz (WoZ) technique \cite{mutlu2012conversational, riek2012wizard}.

\section{Method}

To assess whether different robot heights affect the willingness to comply with a robot's request, we designed an interaction with a robot under two conditions: interaction with a Short robot and interaction with a Tall robot (see Figure \ref{fig:MORPHY - A Modular Robotic Platform}). In the experiment, participants performed a cognitive task on a computer, which was presented as the central part of the experiment. After completing the task, participants were informed that the experiment was over. While waiting to receive their credits, the robotic platform entered the room, carrying a tablet with a call to voluntarily fill out a 300-question questionnaire \cite{soto2017next}. This paradigm is based on previous studies that evaluated compliance \cite{boos2022compliance, cialdini1998social, breckler2006social} and with its definition as "A behavior that was requested by another person or group, i.e., an individual acts in some way because someone else has asked them to do so, while they also had the option to refuse" \cite{breckler2006social}. We evaluated whether the robot's height impacted participants' compliance by measuring the number of questions participants agreed to answer in the different conditions.

 \begin{figure}[t]
     \centering
     \vspace{-0.37cm}
\includegraphics[width=1.2\linewidth]{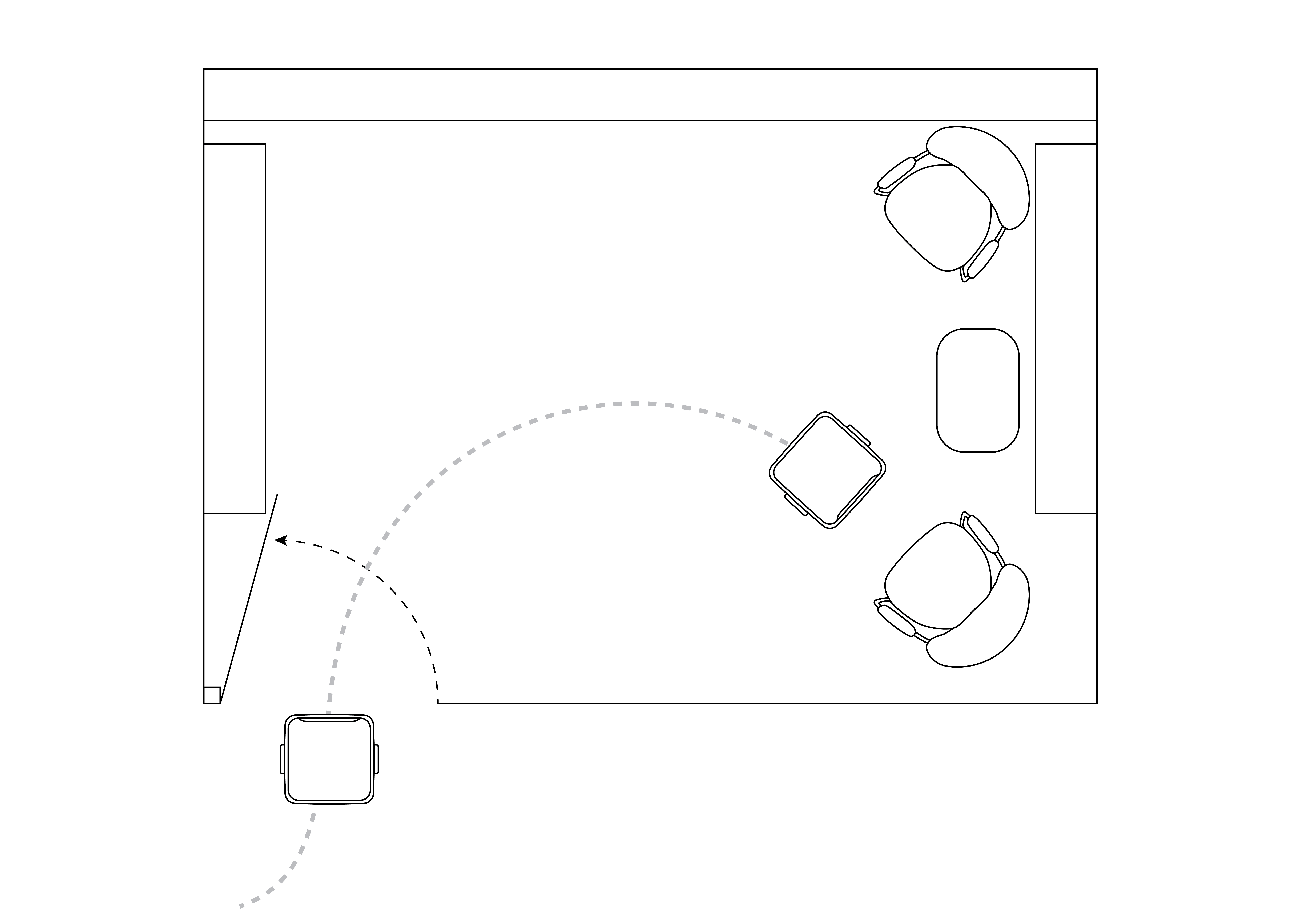}
     \caption{The experiment room settings and the robots' route.}
     \label{fig:ExperimentSettings}
 \end{figure}

\subsection{Participants}
Forty-four undergraduate students from the university participated in the study (31 females, 13 males; Mean age = 23.6, SD = 2.9). All participants completed an informed consent form and received additional course credits for their participation.

\subsection{Experimental settings}
The experiment was carried out in an experimental room, which included two comfortable armchairs and a small side table next to the armchair where the participant would be seated (see Figure \ref{fig:ExperimentSettings}). The participant used the small table to complete the cognitive task on a laptop brought by the researcher. After completing the task, the participant was asked to return the small table to its original position (verifying that the robot could approach the participant directly). On the far side of the room, there was a camera that allowed the researcher to control the robot's movement using the WoZ technique. We decided on a setup in which all participants were seated to reduce variability related to participants' heights.

 \vspace{0.35cm}

\subsection{Experimental design}
\begin{figure}[t]
     \centering
      \vspace{0.3cm}
\includegraphics[width=0.9\linewidth]{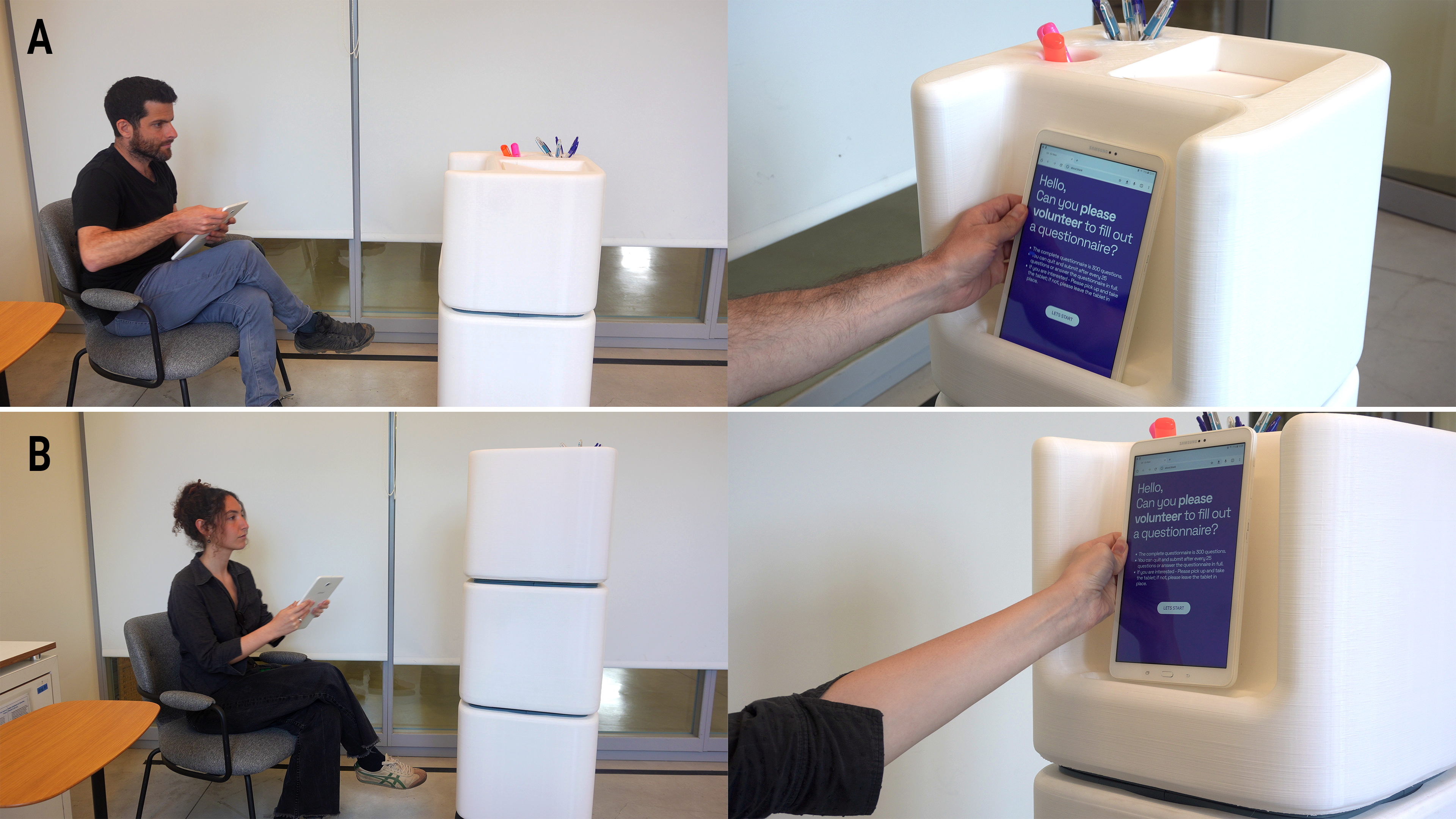}
     \caption{A: Short robot condition; B: Tall robot condition.}
     \label{fig:Experiment}
   \end{figure}

The between-participant experimental design included two conditions: \textit{Short robot} and \textit{Tall robot} (22 \textit{}participants in each condition). The only difference between the conditions was the robot's height (see Figure \ref{fig:Experiment}). The Short robot comprised a base, one body module, and a top module (95 cm in height). The Tall robot comprised a base two body modules and a top module (132 cm in height). To avoid a-priori differences between groups, participants were randomly assigned to the different conditions using a matching technique that balanced gender, tendency for agreeableness \cite{soto2017next}, and Negative Attitudes Toward Robots (NARS) \cite{nomura2006experimental}.

    \vspace{0.35cm}

\subsection{Measures}
Quantitative and qualitative measures were utilized to evaluate the impact of the robot's height on participants' level of compliance and their perception of the robot. 

\subsubsection{Compliance} 
We measured participants' compliance by coding whether participants picked up the tablet with the request to answer 300 questions and how many questions they agreed to answer (as after every 25 questions, they had the option to submit their answers and quit or continue to the next page). 

\subsubsection{The Robotic Social Attributes Scale (RoSAS)}
Following the experiment, participants were asked to complete the Robotic Social Attributes Scale (RoSAS) questionnaire to assess their perception of the robot. The questionnaire is a 9-point Likert scale, ranging from 1 ("low") to 9 ("high"), and comprised of three subscales: warmth, competence, and discomfort \cite{carpinella2017robotic}.   

\subsubsection{Semi-structured interviews}
A semi-structured interview was conducted to map participants’ thoughts and attitudes \cite{boyatzis1998transforming}. Questions focused on participants' experience and thoughts about the robot (e.g., “Describe the experience”, "What did you think about the robot?"). 

\subsection{Procedure}
\begin{figure}[b]
     \centering
\includegraphics[width=0.5\linewidth]{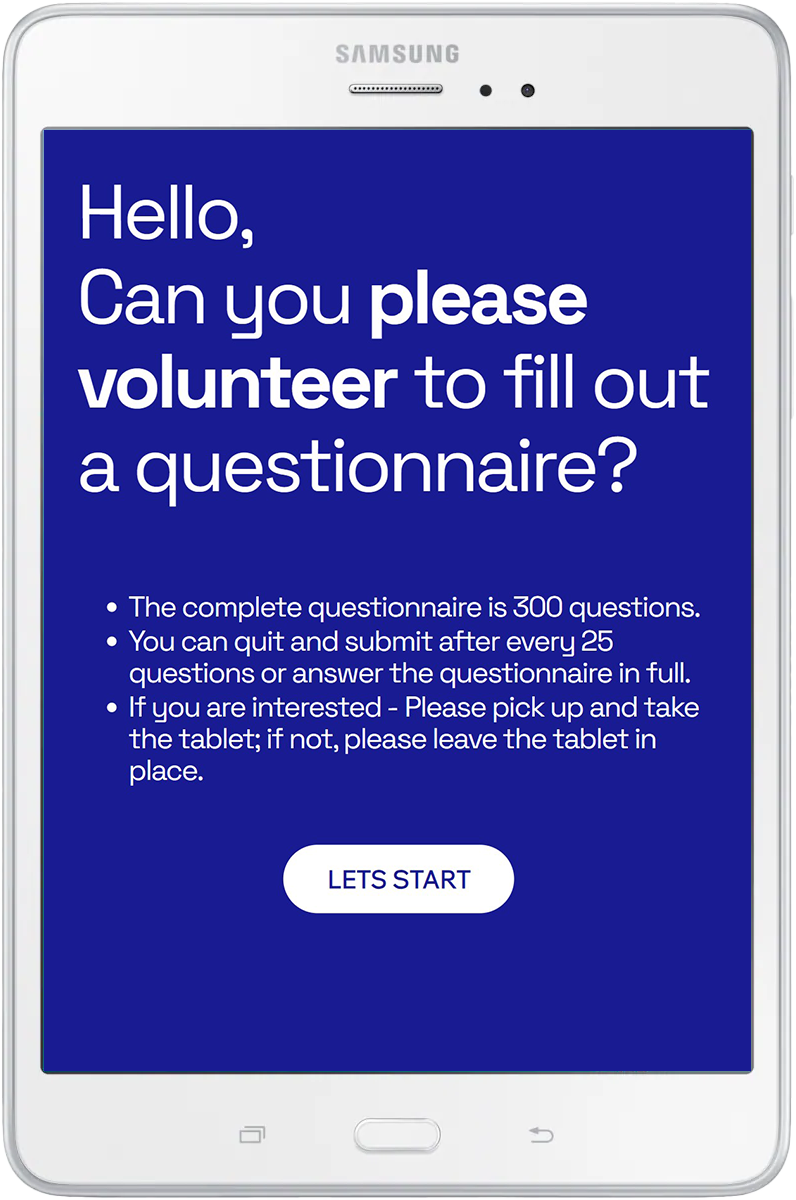}
     \caption{The tablet with the request to complete a long\\ questionnaire voluntarily.}
     \label{fig:Tablet}
 \end{figure}

A few days before the experiment, participants received pre-test questionnaires via email, including a Demographic questionnaire, the Agreeableness Trait scale, and the Negative Attitude towards Robots Scale (NARS). Upon arriving at the lab, participants were informed that the experiment was recorded and that they could withdraw from the experiment at any time without penalty. Before entering the experiment room, the researcher informed the participants that since it is a robotics lab, they might encounter robots as those move freely in all parts of the lab. Participants were then guided into the experiment room and were asked to perform a Stroop task (a simple cognitive task that involves naming the ink color of color words) on a laptop placed on a small table next to them. After completing the Stroop test (approximately 10 minutes), the researcher returned to the room and stated that the experiment was completed. The researcher took the laptop and asked the participant to return the side table to its original position (which cleared a path for a robot to approach the participant). The researcher then explained that the credit approval was in the next room and went to get it. During this time, the robot, based on the experimental condition, entered the room with the tablet containing the BFI-2 300-question long questionnaire (see Figure \ref{fig:Tablet}). The robot approached the participant (who was still seated on the armchair) and stopped at 110 cm from the participant. This distance was validated in a short pilot study as a comfortable distance where the participants could easily notice the invitation to volunteer presented on the Tablet. When the participant picked up the tablet, the robot subtly moved backward to give the participants space until they completed the questionnaire. When the participant returned the tablet to its place on the top part of the robot, the robot backed up and drove outside the experiment room. We note that the robot did not use any communication cues besides its movement and the text on the screen. As the robot exited the room, the researcher returned and asked the participant to complete the RoSAS questionnaire and to participate in a semi-structured interview. The researcher then debriefed the participants, gave them their credits, and made sure they left on a positive note.

\vspace{0.5cm}
\section{Findings}
A Bayesian analysis revealed no early differences between groups in the NARS questionnaire: $BF_{10}=0.06$ and the Agreeableness questionnaire: $BF_{10}=0.2$. The quantitative and qualitative analyses indicated that the robot’s height impacted participants’ compliance with the robot's request.

\subsection{Quantitative analysis}
We initially conducted 2-way ANCOVAs with gender as the additional factor. For simplicity, we report the one-way ANCOVA analyses, as there were no main effects or interactions.

\subsubsection{Compliance}
\begin{figure} [t]
    \centering
    \vspace{-0.1em}
    \includegraphics[width=1\linewidth]{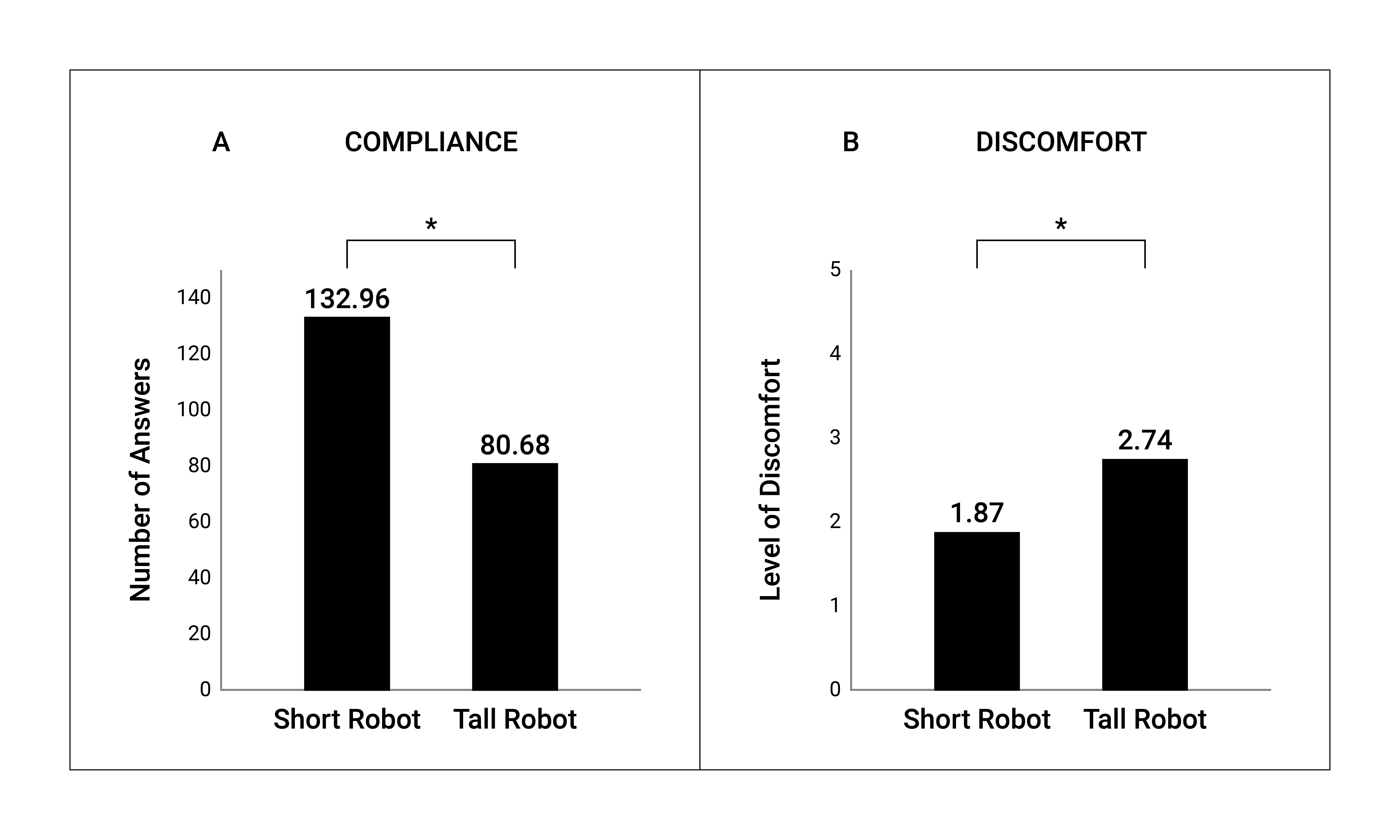}
    \caption{A: Averaged number of questions participants answered; B: Participants' averaged discomfort ratings.}
    \label{fig:Graph}
\end{figure}

Almost all participants in both conditions picked up the tablet (17/22 in the \textit{Short robot} condition; and 15/22 in the \textit{Tall robot} condition; ${\chi}^2$$_\text{(1)}=0.46, p=0.5$). However, a one-way ANCOVA analysis (with participants’ height and age as covariates) revealed that the robot's height had a significant impact on the number of questions participants agreed to answer, F(1, 40) = 4.00, p = 0.05 (significant height covariate, F(1, 40) = 4.00, p = 0.05). The analysis indicated that in the \textit{Short robot} condition, participants agreed to answer more questions (132.9) than in the \textit{Tall robot} condition (80.6), demonstrating higher compliance tendencies (see Figure \ref{fig:Graph}A). 

\subsubsection{Robotic Social Attributes Scale (RoSAS)}
 A one-way ANCOVA analysis (with participants’ height and age as covariates) revealed that the robot's height had a significant impact on participants’ discomfort, F(1, 40) = 3.9, p = 0.05 (no significant covariates). The analysis indicated that participants in the \textit{Short robot} condition experienced less discomfort (1.87) than those in the \textit{Tall robot} condition (2.74; see Figure \ref{fig:Graph}B). The robot's height did not affect the perception of its warmth (\textit{Short robot} - 3.64; \textit{Tall robot} - 3.73) and competence (\textit{Short robot} - 5.37; \textit{Tall robot} - 5.55).

\subsection{Qualitative  analysis - Semi-structured interviews}
The interviews were transcribed and analyzed using Thematic Coding \cite{boyatzis1998transforming}. Three researchers identified initial themes and discussed inconsistencies with a fourth researcher. They then created a list of final themes. Next, they analyzed a subset of the interviews independently and calculated inter-rater reliability (Kappa=81\%). Following the inter-rater reliability, they continued to analyze all interviews. The themes were: volunteering to answer the questionnaire, functional perception of the robot, and general perception of the robot. 

\subsubsection {Volunteering to answer the questionnaire}
Participants in both conditions mentioned their willingness to comply with the robot's request. In both conditions, most participants explained that they complied with the request (\textit{Short robot} - 17/22; \textit{Tall robot} - 15/22). Participants did not attribute their compliance to the robot's characteristics. Instead, they described their behavior as a general tendency for pro-social behavior: \textit{"I could choose... and I just wanted to help."} (P.25, M, \textit{Short robot}); or curiosity \textit{"I don't mind trying new things, and I figured it's a robot with a tablet that's asking questions, so why not"} (P.32, M, \textit{Tall robot}). The few participants who did not volunteer provided practical reasons: \textit{"I did not want to spend more time"} (P.36, F, \textit{Tall robot}).

A similar pattern emerged when participants explained why they did not answer all 300 questions (\textit{Short robot} - 14/22; \textit{Tall robot} - 10/22). They provided general and practical explanations: \textit{"I had an exam and I had to go back to study"} (P.24, F, \textit{Short robot}); \textit{"It was fun at first and then like okay... enough"} (P.34, F, \textit{Tall Robot}).

\subsubsection {Theme 2 - Functional perception of the robot}
Participants in both conditions discussed the robot's functionality. Positive aspects were mentioned more frequently at the \textit{Short robot} condition (16/22) than the \textit{Tall robot} condition (10/22): \textit{"it brings you things, it's cool"} (P.35, F, \textit{Short robot}); \textit{"It would help around with stuff you need… Instead of going and doing stuff yourself, you can use it for specific tasks."} (P.42, F, \textit{Tall robot}). The rest of the participants who discussed the robot's functionality (\textit{Short robot} - 5/22; \textit{Tall robot} - 12/22) provided either neutral or negative descriptions: \textit{"I do not need anything like this unless it could maybe do more things"} (P.43, F, \textit{Short robot}); \textit{"It's too tall and wide, I do not see what people can do with it"} (P.38, F, \textit{Tall robot}).

\subsubsection {Theme 3 - General perception of the robot}
The general perception of the robot followed a similar pattern, indicating a more positive perception of the \textit{Short robot} (14/22) than the \textit{Tall robot} (8/22): \textit{"It was really nice.... very cute... very alive and engaging"} (P.06, M, \textit{Short robot}); \textit{"It felt like a friend"} (P.15, F, \textit{Tall robot}). The rest of the participants who discussed the robot's perception (\textit{Short robot} - 7/22; \textit{Tall robot} - 14/22) provided neutral or negative descriptions: \textit{"It was a little aggressive"} (P.17, F, \textit{Short robot}); \textit{"It made me nervous"} (P.01, F, \textit{Tall robot}).
\section{Discussion}

In this study, we demonstrated that the height of a non-humanoid robotic object can influence participants’ compliance with its request. Our results suggest that interactions with robotic objects evoke social responses, further supporting previous indications that people automatically interpret interactions with highly non-humanoid robots as rich social experiences \cite{erel2019robots, duffy2003anthropomorphism}. However, the direction of the effect that involved a greater tendency to comply with the request when it came from the Short robot, was opposite to the one typically observed in human-human interactions, where compliance is higher for requests coming from taller people due to their increased persuasiveness and authority \cite{young1996height, higham1992rise, stogdill1948personal}. Our findings revealed that when the request came from a Short robot, participants agreed to answer significantly more questions in an additional questionnaire than when the request came from a Tall robot. Our findings, therefore, challenge the common belief that interactions with robots are similar to interactions between humans \cite{nass1994computers, erel2024rosi}. We argue that it is essential to map the social impact of specific robotic features (such as height) when designing different robotic morphologies, as it cannot be applied or assumed based on the vast knowledge in social psychology (which has already mapped similar impacts in human-human interactions). 

Participants' descriptions and perceptions of the Short robot may provide insight into their increased compliance with its request. Although participants did not explicitly mention the robot’s height, they often used positive terms to describe it, stating that it was \textit{"Cute"}, \textit{"Nice"}, \textit{"Cool"}, and \textit{"Engaging"}. The Tall robot was described more frequently by neutral or negative terms, such as \textit{"Normal"}, \textit{"Harmless"}, \textit{"Weird"}, and \textit{"Scary"}. This explanation was supported by the RoSAS data, indicating increased discomfort when interacting with the taller robot. This suggests that the height of a robotic object may influence the valence of its perception, which in turn determines the level of its persuasiveness, with a positive perception leading to increased compliance. Another possible explanation may be attributed to the perception of the shorter robot as more vulnerable or child-like. Although there are no indications of this possibility in our results, it warrants further investigation. 

More broadly, our findings imply that people's compliance with robotic objects' requests may be shaped by social dynamics that are distinct from those in human-human interaction. While compliance with human requests is associated with the person's power, compliance with a robotic object's request may be based on simpler processes related to the robot’s perception and perceived "cuteness". Hence, mapping robotic characteristics that would enhance its perception and minimize discomfort may be sufficient for significantly increasing people's compliance with its requests. 

Taken together, our findings indicate the importance of mapping and carefully considering robotic design characteristics when designing robotic objects. We show that the robot's characteristics can significantly affect its social impact. The nature of the impact is shaped by processes unique to the interactions with robots. It is, therefore, important to further understand and map these processes and the social influence of different robotic design characteristics.

\section{Limitations}
We used a robotic object designed as a service table, which limits generalization. The effect of height should be further examined across different robotic morphologies. Second, the interaction was brief and simple. Factors such as the interaction’s content, emotional tone, and the robot’s functionality may also influence its social outcomes. Future studies should also explore additional design characteristics (e.g., color, form, material), broader height variations, and the impact of the robot's relative height. Additionally, we did not examine whether the effect persists over time, highlighting the need to test for potential habituation. Lastly, we followed a strict protocol to mitigate influences related to “good subject effect” in the interviews \cite{nichols2008good}. 
\section{Conclusion}
We examined the impact of a non-humanoid robotic object's height on participants' compliance with the robot's request. As in human-human interactions, our results show that a robot’s height can have a significant social impact. However, when the robot lacks a humanoid form, this effect may follow a pattern opposite to that observed in human-human interactions. We highlight the need for robot designers and practitioners to consider their robots' characteristics and how these shape their social impact. 
\section{Acknowledgment}
We thank Ziv Keidar, May Bini, Noa Rivka Kirtchuk, Zohar Fein, Hila Zohar and Srivatsan Chakravarthi Kumaran for their great contribution to this work.






\bibliography{BIBMAIN}
\bibliographystyle{IEEEtran}

\end{document}